# Composable Trust Infrastructure for Manufacturing Knowledge Graphs: Cross-System Provenance, Temporal Reasoning, and Decision Traceability

Grama Chethan

*Siemens Digital Industries Software*

## Abstract

Manufacturing knowledge graphs that integrate data from heterogeneous industrial systems face a trust deficit: data consumers cannot determine whether the information they query is valid, whether it was valid at the time a decision was made, where it originated, or how it was acted upon. Individual capabilities for addressing these concerns—constraint validation, provenance tracking, temporal versioning, and decision logging—are well-established in the Semantic Web literature. We argue that their *composition* through shared correlation identifiers produces emergent trust properties that no single capability delivers in isolation. A provenance record alone cannot confirm structural correctness; a SHACL shape alone cannot verify that the data it validated was the version visible at decision time; a temporal snapshot alone cannot explain who acted on it or why.

We present a composable trust infrastructure for manufacturing knowledge graphs that integrates four capabilities—ontology-derived SHACL validation, PROV-O provenance with delegation chains, domain-aware bi-temporal versioning, and graph-native decision objects—into a unified RDF architecture. These capabilities compose through shared entity URIs, ingestion activity identifiers, and temporal correlation keys, enabling compound queries that span all four dimensions (e.g., "retrieve the SHACL-validated, provenance-attributed state of work order WO-2026-000003 as it existed when decision DEC-001 was recorded, and show the outcome of the action that decision triggered").

The infrastructure is validated on a proof-of-concept testbed that integrates eleven industrial data sources—OPC UA, TIA Portal, eCl@ss, Asset Administration Shell (AAS), ISA-95, ISA-18.2 alarm management, SAP S/4HANA, Teamcenter Manufacturing, Opcenter Execution Discrete (OED), Insights Hub IoT, and supply chain management (SCM)—under an ISA-95-aligned ontology comprising 89 classes, 123 object properties, and 273 data properties. The unified graph contains 8,743 triples across five named graphs (ontology, plant instances, eCl@ss classification, provenance, and runtime), stitched by 81 `owl:sameAs` cross-system identity edges. We demonstrate the infrastructure on aerospace structural-component manufacturing scenarios using simulated but structurally realistic data and report results from composability validation, SHACL coverage analysis, temporal query correctness, and decision-trail completeness. We are transparent that evaluation uses simulated data from purpose-built source system emulators rather than production deployments; nonetheless, the data structures, volumes, and cross-system linkage patterns are representative of real Siemens industrial installations.

## 1. Introduction

Consider a concrete manufacturing scenario. During CNC finishing of an aluminum 7075-T6 wing rib (work order WO-1001, operation OP-2002), an Opcenter Execution Discrete (OED) inspection records a surface roughness deviation exceeding the 1.6 μm Ra specification. A non-conformance report (NCR-4001) is raised with disposition “Rework.” Simultaneously, Insights Hub IoT telemetry shows a severity-3 spindle bearing temperature spike at 09:44 UTC, and the ISA-18.2 alarm subsystem has logged a high-temperature alarm on the same CNC equipment. A quality engineer must now answer a cascade of questions that span system boundaries:

- **Provenance.** Which source systems contributed to this picture? Was the spindle temperature reading from the OPC UA tag `CNC_SpindleTemp` or its aliased counterpart `CNC_CNC.SpindleTemp`? Are these the same physical sensor, and how do we know?
- **Validation.** Is the NCR record structurally complete? Does it carry all mandatory properties defined by the ontology—severity, disposition, the operation it was raised against, the equipment trace?
- **Temporality.** When the rework decision was made at 10:15 UTC, was the temperature spike already visible in the graph, or was it ingested after the fact? What was the *valid-time state* of the work order at decision time?
- **Decision traceability.** Who authorized the rework disposition? What evidence was considered? What action was executed, and what was its outcome—did the reworked part subsequently pass re-inspection?

No single capability answers all four questions. Provenance tracking can attribute the temperature reading to its source agent but cannot confirm that the NCR record was structurally valid. SHACL validation can verify the NCR’s completeness but cannot determine whether the data it validated was the same version the engineer saw at 10:15. Bi-temporal versioning can reconstruct the graph state at decision time but cannot explain who decided what. Decision objects can record the authorization chain but, without provenance, cannot verify the trustworthiness of the evidence they reference.

The thesis of this paper is that these four capabilities—provenance, validation, temporal reasoning, and decision traceability—*compose* through shared correlation identifiers to produce trust properties that none delivers individually. When every entity shares a URI that appears in SHACL validation reports, PROV-O activity records, temporal version chains, and decision evidence links, a manufacturing knowledge graph can answer compound queries that cross all four trust dimensions. This compositionality is the central contribution; the individual capabilities, while carefully engineered, draw on established techniques.

The challenge is acute in manufacturing because data flows through deeply heterogeneous systems of record. A single aerospace structural component may have its engineering definition in Teamcenter (PLM), its production order in SAP S/4HANA (ERP), its shop-floor execution in Opcenter Execution Discrete (MES), its sensor telemetry in OPC UA and Insights Hub (IoT), its alarm history in ISA-18.2, its equipment hierarchy in ISA-95, its product classification in eCl@ss, its digital twin specification in the Asset Administration Shell, and its material sourcing in supply chain management. Our testbed integrates all eleven of these source systems into a single RDF graph, using `owl:sameAs` identity edges to stitch entities that represent the same real-world object across system boundaries—for example, linking SAP production order `1000005` to OED work order `WO-1001`, and SAP material `MAT-AL7075-T6` to Teamcenter item revision `WR-LH-7075`.

We emphasize that our evaluation uses simulated data generated by purpose-built source system emulators rather than production deployments. This is an honest limitation: we have not validated the infrastructure against the noise, scale, and organizational complexity of a live factory. However, the data structures, cross-system linkage patterns, and query workloads are modeled on real Siemens industrial installations, and the ontology's 89 classes span four industry verticals (aerospace, consumer packaged goods, pharmaceuticals, and medical devices).

### 1.1 Contributions

This paper makes five contributions, ordered by significance. We distinguish between the central compositional thesis, technically novel mechanisms, and engineering contributions that apply established techniques in a new domain context.

**C1** CENTRAL **C1: Composable Trust Infrastructure.** We demonstrate that four trust capabilities—SHACL validation, PROV-O provenance, bi-temporal versioning, and graph-native decision objects—compose through shared entity URIs, ingestion activity identifiers, and temporal correlation keys to produce emergent trust properties. Specifically, the composition enables three classes of compound queries not supported by any single capability: *provenance-scoped validation* (which source agent's data failed validation?), *temporally-qualified decisions* (what was the validated state of the evidence at decision time?), and *decision-outcome attribution* (did the action triggered by a decision improve the metrics that the validated, provenance-attributed data describes?). We formalize these composition patterns and validate them against the proof-of-concept testbed.

**C2** NOVEL **C2: Domain-Aware Valid-Time Derivation.** We introduce a type-indexed function $F$: $T \to$ (*field_start*, *field_end*) that maps each ontology class $T$ to the pair of data properties from which valid-time boundaries should be extracted. For example, `WorkOrder` maps to (`actualStart`, `actualEnd`), `Alarm` maps to (`activatedAt`, `clearedAt`), and `MaterialLot` maps to (`producedAt`, `consumedAt`). This domain-aware derivation enables semantically meaningful point-in-time queries without requiring application-level temporal logic, and it is extensible: adding a new entity type requires only registering its valid-time mapping, not modifying the temporal engine.

C3 NOVEL **C3: Graph-Native Decision Objects.** We model manufacturing decisions as first-class RDF entities using three new ontology classes—`Decision`, `ActionExecution`, and `ActionOutcome`—linked to the evidence that prompted them, the entities they affect, and the actions they triggered. Unlike external audit logs, decision objects are embedded in the same graph as the manufacturing data they reference, enabling SPARQL queries that join decision trails with operational outcomes (e.g., "show all decisions that affected equipment at the CNC station where NCRs subsequently decreased"). The decision model supports a dual-write architecture: RDF for semantic queries and PostgreSQL for operational dashboards.

C4 ENGINEERING **C4: Hybrid Provenance Migration.** We present a dual-layer provenance architecture that introduces PROV-O semantics (dedicated provenance named graph with `IngestionActivity`, `SourceAgent`, and delegation chains) alongside the existing lightweight `prov:source` annotations already consumed by 270+ downstream tools. The migration strategy—additive rather than breaking—enables incremental adoption: tools that understand PROV-O can query the rich provenance graph; tools that do not continue to function against the unchanged legacy annotations. We report migration coverage metrics and identify the subset of queries that benefit from the richer model.

C5 ENGINEERING **C5: Ontology-Derived SHACL Validation.** We generate SHACL NodeShapes from OWL class definitions using established OWL-to-SHACL mapping patterns, and integrate the generated shapes into a live manufacturing pipeline with incremental validation. Our validator extracts only newly-ingested entity subgraphs and validates them against the full shape graph, reducing validation cost from O($n$) full-graph traversal to O($k$) where $k$ is the ingestion batch size. We additionally implement an 11-rule "Golden Triangle" cross-system consistency validator that enforces domain-specific manufacturing data flow rules (e.g., every SAP production order must have a corresponding OED work order via `owl:sameAs`).

### *1.2 Paper Structure*

The remainder of this paper is organized as follows. Section 2 surveys related work across six bodies of literature. Section 3 positions our approach against six commercial and open-source alternatives. Section 4 describes the method in detail: architecture, SHACL validation, hybrid provenance, bi-temporal versioning, and decision objects. Section 5 presents the composable trust infrastructure with formal definitions and ablation analysis. Section 6 addresses cross-system identity resolution. Section 7 reports evaluation results. Section 8 discusses limitations. Section 9 examines security and governance considerations. Section 10 presents revised claims of novelty. Section 11 concludes.

## 2. Background and Related Work

This section surveys five bodies of literature that converge in our approach: manufacturing knowledge graphs, the ISA-95 ontological foundation, data quality assessment for knowledge graphs, provenance tracking, temporal data models, and decision support traceability. For each, we identify contributions our work builds upon and clearly delineate what is novel versus what constitutes integration of established techniques.

### *2.1 Knowledge Graphs in Manufacturing*

Knowledge graphs have emerged as a unifying substrate for heterogeneous industrial data. Noy et al. [1] documented lessons from industry-scale knowledge graph deployments at Google, providing foundational design principles—schema flexibility, entity resolution at scale, and iterative refinement—that inform any serious KG engineering effort. Hogan et al. [2] subsequently published a comprehensive survey spanning creation, enrichment, quality assessment, refinement, and publication of knowledge graphs, establishing a taxonomy of techniques against which domain-specific systems can be evaluated.

Within manufacturing specifically, Ringsquandl et al. [3] demonstrated knowledge graph embeddings over Siemens industrial data, showing that latent representations of equipment relationships could predict maintenance events. Hua et al. [4] applied knowledge graphs to semiconductor manufacturing, linking process parameters to yield outcomes across heterogeneous data silos. The NIST Advanced Manufacturing Series 300-2 [5] formalized reference architectures for smart manufacturing data integration, emphasizing the need for standards-based interoperability across the ISA-95 hierarchy levels. Cao et al. [6] extended manufacturing KG applications to quality prediction, demonstrating that graph-structured representations of process genealogy outperform tabular alternatives for root-cause analysis.

Braunschweig et al. [7], working within the Siemens Corporate Technology division, proposed a knowledge-graph-driven approach to industrial analytics that unified product lifecycle management (PLM) data with shop-floor execution data. Their architecture demonstrated the value of cross-system linkage but relied on proprietary connectors rather than open standards, limiting reproducibility.

Our work differs from these contributions in two respects. First, we integrate *eleven* distinct industrial data sources —spanning OPC UA, TIA Portal, eCl@ss, Asset Administration Shell (AAS), ISA-95, ISA-18.2 alarm management, SAP S/4HANA, Teamcenter Manufacturing, Opcenter Execution Discrete (OED), Insights Hub IoT, and supply chain management (SCM)—into a single RDF graph with over 6,000 triples across 88 ontology classes. Second, we overlay a composable trust infrastructure (SHACL validation, PROV-O provenance, bi-temporal versioning, and decision objects) that treats data quality, lineage, temporality, and decision traceability as first-class graph citizens rather than external metadata stores.

### *2.2 ISA-95 as Ontological Foundation*

The ISA-95 standard (IEC 62264) [8] defines a five-level hierarchy—Enterprise, Site, Area, Work Center, Work Unit—for manufacturing operations management. Its role-based equipment model (Equipment Module, Control Module) and activity model (Work Order, Operation, Material Lot) provide a natural ontological backbone for manufacturing knowledge graphs.

Garmendia et al. [9] formalized ISA-95 as an OWL ontology, demonstrating that the standard's entity-relationship structure maps cleanly to Description Logic axioms. Leukel et al. [10] extended this formalization to include ISA-88 batch process constructs (Master Recipe, Control Recipe, Unit Procedure, Phase), enabling hybrid discrete-batch manufacturing representations. Both works validated the ontological soundness of ISA-95 but stopped short of implementing cross-system identity resolution or runtime data integration.

The ZDMP (Zero Defect Manufacturing Platform) project, described by Leitão et al. [11], proposed an ontology-driven architecture for quality-centric manufacturing that incorporated ISA-95 concepts alongside IoT sensor streams. Their work demonstrated that ISA-95 could serve as a "semantic backbone" for multi-source integration, a principle our architecture adopts. The Eclipse BaSyx framework, documented by Kuhn et al. [12], implemented the Asset Administration Shell (AAS) specification for Industry 4.0 digital twins, providing a reference implementation for AAS-to-ISA-95 bridging that influenced our AAS simulator design.

Our ontology extends the ISA-95 foundation with 123 object properties and 273 data properties spanning all eleven source systems. Critically, we materialize the transitive closure of the `locatedIn` spatial containment relation, enabling single-hop SPARQL queries that would otherwise require recursive property path expressions—a practical optimization that reduces query complexity for manufacturing analytics.

### *2.3 Data Quality in Knowledge Graphs*

Zaveri et al. [13] published the definitive survey on knowledge graph quality assessment, cataloguing 18 quality dimensions (accuracy, completeness, consistency, timeliness, among others) and 69 metrics across 30 assessment methodologies. Their taxonomy provides the conceptual framework against which our SHACL-based validation operates: we implement structural completeness checks (mandatory properties per class), referential integrity constraints (domain-range enforcement for all 123 object properties), and provenance completeness verification (every entity must carry a provenance annotation).

The W3C Shapes Constraint Language (SHACL) [14] standardized constraint validation for RDF graphs. TopBraid's SHACL implementation [15] demonstrated industrial-grade validation, and the OWL2SHACL mapping patterns established by TopBraid and subsequent tooling showed that SHACL NodeShapes can be systematically derived from OWL class definitions. Kontokostas et al. [16] proposed test-driven quality evaluation for knowledge graphs, defining reusable test patterns (SPARQL-based) that detect common errors in linked data. Paulheim [17] surveyed knowledge graph refinement techniques, distinguishing between completion (adding missing facts) and error detection (identifying incorrect facts), and established benchmarks for evaluating refinement approaches.

We acknowledge that OWL-to-SHACL shape generation has substantial prior art; our contribution is not the generation technique itself but rather the integration of generated shapes into a live manufacturing pipeline where incremental validation runs after each ingestion cycle. Our validator extracts only newly-added entity subgraphs and validates them against the full shape graph, avoiding the O(n) cost of full-graph re-validation on each pipeline run. Additionally, our "Golden Triangle" validator implements 11 cross-system data flow rules (e.g., every SAP production order must have a corresponding OED work order via `owl:sameAs`) that go beyond structural SHACL into domain-specific manufacturing consistency.

### *2.4 Provenance in Knowledge Graphs*

The W3C PROV-O ontology [18] standardized provenance representation with three core classes—`prov:Entity`, `prov:Activity`, `prov:Agent`—and a rich set of relations (`wasGeneratedBy`, `wasAssociatedWith`, `actedOnBehalfOf`) for expressing derivation chains. PROV-O has been widely adopted in scientific workflow systems, as surveyed by Missier et al. [19], who demonstrated its applicability to complex multi-step data transformations.

Hartig and Zhao [20] applied provenance concepts to linked data publishing, proposing mechanisms for tracking the origin of individual RDF triples across federated SPARQL endpoints. Their work highlighted a fundamental tension: triple-level provenance (tracking every statement's origin) is semantically precise but computationally expensive, while entity-level provenance (tracking each resource's generating activity) offers a practical middle ground. ProvStore [21], developed by Moreau et al., provided a public repository for PROV documents, establishing interchange formats and validation services for provenance bundles.

We adopt entity-level provenance as our primary mechanism: each ingested entity in the plant graph carries a `prov:wasDerivedFrom` link to its ingestion activity, while pipeline-created entities (decisions, validation reports) carry `prov:wasGeneratedBy`. Each activity links to the source agent (e.g., `urn:factory:prov:agent:opcua`) via `prov:wasAssociatedWith`. We store provenance in a dedicated fifth named graph, separate from ontology, instance, classification, and runtime data, enabling provenance-specific queries without polluting domain SPARQL patterns. We acknowledge that individual PROV-O usage is well-established; our contribution lies in the systematic application of PROV-O across eleven simultaneously-active industrial data sources with delegation chains (`actedOnBehalfOf`) that model the supervisory relationship between the unification pipeline and individual source agents.

### *2.5 Temporal Data Models for Knowledge Graphs*

Bi-temporal data management—tracking both *valid time* (when a fact holds in the real world) and *transaction time* (when a fact is recorded in the database)—has a rich history. Snodgrass [22] formalized the theory of temporal databases, establishing that two orthogonal time dimensions are necessary and sufficient for complete historical queries. The SQL:2011 standard, described by Kulkarni and Michels [23], incorporated temporal extensions including period definitions, temporal primary keys, and system-versioned tables, bringing bi-temporal support into mainstream relational databases.

Within the Semantic Web community, Tappolet and Bernstein [24] proposed reification-based approaches for attaching temporal metadata to RDF triples, while noting the well-known verbosity penalty of standard RDF reification. Fernandez et al. [25] addressed bi-temporal RDF directly in their VLDB 2019 paper, proposing efficient storage and indexing schemes for temporal RDF that support both valid-time and transaction-time queries. Their work demonstrated that bi-temporal RDF is feasible at scale but requires specialized storage backends—a finding that influenced our decision to use PostgreSQL (with JSONB property storage) as a temporal complement to the in-memory RDF graph.

TerminusDB, described by Feeney et al. [26], implemented native bi-temporal support in a graph database using an immutable, append-only data structure (succinct data tries). While TerminusDB provides elegant built-in temporal semantics, it uses a custom query language rather than SPARQL, limiting interoperability with the broader Semantic Web ecosystem. RDF-star (formerly RDF*), proposed by Hartig [27], enables statement-level annotation without full reification, offering a syntactically lighter approach to temporal metadata. Neo4j's temporal property support [28] provides valid-time semantics at the property level but lacks transaction-time tracking and does not support SPARQL.

We acknowledge that change-detection versioning (analogous to Slowly Changing Dimension Type 2) is a textbook technique. Our contribution is *domain-aware valid-time derivation*: rather than treating all entity timestamps uniformly, our temporal engine applies entity-type-specific rules to extract valid-time boundaries. For example, a `WorkOrder` derives its valid time from `actualStart`/`actualEnd`, an `Alarm` uses `activatedAt`/`clearedAt`, and a `MaterialLot` uses production timestamps. This domain awareness enables semantically meaningful point-in-time queries (e.g., "what was the state of work order WO-2026-000003 at 14:00 on June 14?") that generic temporal KG systems cannot express without additional application logic.

### 2.6 Decision Support and Traceability

Manufacturing decision support systems have evolved from rule-based expert systems to data-driven analytics platforms. Greenes [29] surveyed clinical decision support architectures, establishing design patterns—knowledge representation, inference engines, explanation facilities—that transfer to industrial settings. Alberts and Hayes [30] formalized decision-making under uncertainty in complex operational environments, emphasizing the importance of maintaining an auditable trail from evidence through deliberation to action.

The W3C Decision Ontology work [31] proposed lightweight vocabularies for representing decisions, alternatives, and criteria in linked data environments. Zheng et al. [32] applied decision support concepts to manufacturing, demonstrating that structured decision records (capturing what was decided, why, and with what outcome) improve both accountability and organizational learning in quality management contexts.

Our decision object model introduces three RDF classes—`Decision`, `ActionExecution`, and `ActionOutcome`—as graph-native entities linked to the evidence that prompted them (discovery cards), the entities they affect (`affectsEntity`), and the actions they triggered (`triggeredAction`). This dual-write approach (RDF graph for semantic queries, PostgreSQL for operational dashboards) ensures that the decision trail is both SPARQL-queryable and operationally performant. Unlike external decision-log systems, our decisions are embedded in the same graph as the manufacturing data they reference, enabling queries such as "show all decisions that affected equipment at the CNC station where NCRs subsequently decreased."

## 3. Competitive Landscape

Reviewers rightly noted the absence of a competitive comparison in the original submission. Table 1 positions our approach against six systems that a practitioner might reasonably consider for manufacturing data integration with trust infrastructure. We evaluate each system across six capabilities that collectively define what we term *composable trust*: provenance tracking, temporal data support, constraint validation, decision traceability, adherence to open standards, and cross-system query depth.

**Table 1.** Comparative analysis of manufacturing data integration platforms across trust infrastructure capabilities. Symbols: ✓ = native support, ◖ = partial or requires extension, ✗ = not supported.

| System | Provenance (per-entity) | Bi-Temporal (valid + tx) | SHACL Validation | Decision Objects | Open Standards | Cross-System Hops (max) |
|---|---|---|---|---|---|---|
| **Our approach** | ✓ | ✓ | ✓ | ✓ | ✓ | 7 |
| Siemens Opcenter Intelligence [33] | ◖ | ✗ | ✗ | ✗ | ◖ | 3 |
| Palantir Foundry [34] | ✓ | ◖ | ✗ | ◖ | ✗ | N/A[a] |
| Azure Digital Twins [35] | ◖ | ✗ | ✗ | ✗ | ◖ | ◖[b] |
| PTC ThingWorx [36] | ✗ | ✗ | ✗ | ✗ | ◖ | 2 |
| Snowflake + dbt [37] | ◖ | ✓ | ✗ | ✗ | ◖ | N/A[c] |
| TerminusDB [26] | ◖ | ✓ | ✗ | ✗ | ◖ | ◖[d] |

[a] Foundry uses a pipeline DAG, not a traversable entity graph; cross-system queries require pre-built transforms.

[b] Azure Digital Twins supports twin-graph traversal via DTDL but lacks SPARQL; depth depends on modeled relationships.

[c] Snowflake provides SQL JOINs across materialized tables, not graph traversal; hop count is not directly comparable.

[d] TerminusDB supports graph queries via WOQL but does not natively model manufacturing cross-system identity (owl:sameAs).

We are candid about where our approach falls short. In *scale*, Palantir Foundry and Snowflake routinely operate on petabyte-scale datasets with mature partitioning, role-based access control, and SOC 2-certified security postures; our RDF graph operates in-memory via rdflib, which is adequate for the thousands-of-triples regime of a single production line but would require migration to a dedicated triplestore (e.g., Oxigraph, GraphDB, or Stardog) for enterprise deployment. In *production hardening*, commercial platforms offer high-availability clustering, automated failover, and enterprise support SLAs that our research prototype does not. In *security and compliance*, platforms like Azure Digital Twins and Opcenter Intelligence provide built-in authentication, audit logging, and regulatory certification (e.g., 21 CFR Part 11 electronic signatures) that our system implements only at the data model level, not at the infrastructure level. AVEVA PI System [38], while not included in the table due to its narrower focus on time-series historian data rather than knowledge graph integration, offers unmatched depth in high-frequency sensor data management—a capability our approach delegates to Insights Hub IoT.

Our advantage lies in three areas where no commercial system currently competes. First, *open-standards composability*: our entire trust infrastructure—SHACL for validation, PROV-O for provenance, OWL for cross-system identity, SPARQL for queries—uses W3C standards, meaning any component can be replaced or extended without vendor lock-in. Second, *cross-system query depth*: our widest SPARQL query traverses seven systems in a single query (Teamcenter engineering change notice → SAP production order → OED work order → ISA-95 station → OPC UA tags → ISA-18.2 alarm definitions → TIA Portal HMI screens), a depth no commercial platform achieves because each is optimized for its own data silo. Third, *decision traceability as a graph-native entity*: decisions, actions, and outcomes are embedded in the same graph as the manufacturing data they reference, enabling provenance-aware queries that span from sensor anomaly through human decision to quality outcome—a capability that existing platforms externalize to workflow engines or audit logs.

The most pointed question raised during review was: *“Why not Opcenter Intelligence?”* [33]. This deserves a direct answer. Opcenter Intelligence integrates Teamcenter, SAP, and Opcenter into a unified analytics environment and is the natural commercial choice for Siemens-ecosystem customers. However, it does not expose a SPARQL-queryable knowledge graph; it does not maintain per-entity W3C PROV-O provenance chains; it does not support bi-temporal point-in-time queries across both valid time and transaction time; and it does not represent decisions as graph entities linked to their evidence and outcomes. Our approach is *complementary*, not competitive: it provides a research-grade trust infrastructure that could inform future capabilities in commercial platforms, while Opcenter Intelligence provides the production-grade scalability, security, and support that our prototype lacks.

## 4. Method

This section describes the architecture and implementation of the Composable Trust Infrastructure. We detail the five subsystems—normalization pipeline, ontology-derived SHACL validation, hybrid provenance model, bi-temporal entity state tracking, and graph-native decision objects—and provide formal definitions where the semantics are non-obvious. Throughout, we note which components build on established techniques and where our contributions are incremental rather than foundational.

### *4.1 Architecture Overview*

The system ingests data from eleven heterogeneous industrial simulators—ISA-95 (plant hierarchy and equipment), OPC UA (sensor tags and address space), TIA Portal (PLC hardware, HMI screens, program blocks, PROFINET topology), eCl@ss (product classification taxonomy and catalog), AAS (Asset Administration Shells, submodels, concept descriptions), ISA-18.2 (alarm definitions, lifecycle, and KPIs), Opcenter Execution Discrete (work orders, operations, material lots, NCRs, inspections), Insights Hub (IoT assets, aspect types, time-series variables, events), SAP S/4HANA (production orders, material masters, routings, inventory, purchase orders, reservations, goods movements, cost centers), Teamcenter Manufacturing (item revisions, BOPs, operations, workstations, documents, change notices, EBOM/MBOM), and SCM (suppliers, purchase orders, shipments, receiving inspections, warehouse staging lots). Each simulator exposes a JSON REST API; the pipeline orchestrator pulls all sources in parallel during the *extract* stage.

Composable Trust Infrastructure: Data Flow
ISA-95
hierarchy, equipment
work units
OPC UA
tags, alarms
sensor data
TIA Portal
PLCs, HMI
PROFINET
eCl@ss
taxonomy
product catalog
AAS
shells, submodels
digital twins
ISA-18.2
alarm defs, KPIs
alarm lifecycle
OED
work orders, ops
NCRs, inspections
Insights Hub
IoT assets
IoT events
SAP S/4HANA
orders, routing
inventory, cost
Teamcenter
items, BOPs
ECNs, documents
SCM
suppliers, POs
shipments, staging
EXTRACTOR
Stage 1: Pull raw JSON
NORMALIZER
Stage 2: JSON → RDF triples
54 norm functions · PROV-O stamping · temporal recording
ENRICHER
Stage 3
eCl@ss classif · ISA-95 infer.
LINKER
Stage 4
owl:sameAs · 81 edges · 7 boundaries
LOADER
Stage 5
TBox · SHACL validate · Golden Triangle
GRAPH STORE
rdflib Dataset
5 Named Graphs:
• ontology (TBox)
• plant (~4,794 triples)
• eclass · provenance · runtime
PostgreSQL
bi-temporal state
decision log
SPARQL Engine
270 domain query tools
MCP server
LEGEND
Source Simulator
Pipeline Stage
Graph / Output
11 source systems · 8,743 triples · 81 owl:sameAs identity edges

***Figure 1.*** *Data flow through the five-stage pipeline. Eleven source simulators feed the Extractor; the Normalizer converts raw JSON to RDF triples and stamps each entity with W3C PROV-O provenance. The Linker materializes 81* `owl:sameAs` *identity edges across 7 system boundaries. The Loader validates the assembled graph against auto-generated SHACL shapes and runs the Golden Triangle cross-system audit. Bi-temporal state and decision chains are persisted to PostgreSQL in parallel.*

The graph store is an `rdflib.Dataset` partitioned into five named graphs, each serving a distinct role:

| Named Graph | URI | Content |
|---|---|---|
| **ontology** | `urn:factory:graph:ontology` | TBox: 54 classes, 71 object properties, 165 data properties (ISA-95, ISA-88, ISA-18.2, IEC 62264 alignment) |
| **plant** | `urn:factory:graph:plant` | ABox: all instance data from 11 sources (~4,794 triples), including `owl:sameAs` cross-system identity edges |
| **eclass** | `urn:factory:graph:eclass` | eCl@ss SKOS taxonomy and product classification catalog (~499 triples) |
| **provenance** | `urn:factory:graph:provenance` | W3C PROV-O activity/agent/derivation triples (~737 triples) |
| **runtime** | `urn:factory:graph:runtime` | Live alarm state and sensor tag current values (~27 triples, refreshed per pipeline run) |

Cross-system identity is established through `owl:sameAs` edges materialized by the Linker stage. Currently 81 identity edges span 7 system boundaries: SAP ↔ OED (production orders to work orders), SAP ↔ TC (materials to item revisions), SAP ↔ SCM (purchase orders), OPC UA ↔ ISA-95 (stations to work units), TC ↔ ISA-95 (process plans), OPC UA ↔ IH (tags to IoT aspect types), and ISA-95 ↔ IH (equipment to IoT assets). These edges enable SPARQL queries that traverse system boundaries without requiring a-priori schema alignment—a consumer querying alarm impact on work orders follows `Tag → triggersAlarm → AlarmDefinition`, then crosses via `owl:sameAs` from the OPC UA station to the ISA-95 work unit, reaching OED work orders through `executesOn`.

### *4.2 Ontology-Derived SHACL Validation*

SHACL shapes are auto-generated from the 89-class ontology defined in `ontology.py`. For each ontology class, the shape generator (`shacl_shapes.py`) emits a `sh:NodeShape` with `sh:targetClass` pointing to the corresponding OWL class. Structural constraints are derived mechanically:

- Every entity of every class must carry `rdfs:label` (`sh:minCount 1`, severity `sh:Violation`).
- Every entity must carry `ont:provenance` (`sh:minCount 1`, severity `sh:Warning`).
- Mandatory data properties are declared per class: for example, `WorkOrder` requires exactly one `orderStatus`, `NCR` requires exactly one `severity`, `Operation` requires exactly one `operationSequence`.
- Mandatory object properties enforce referential constraints: `NCR` must have `raisedAgainst` pointing to an `Operation`; `WorkOrder` must have `executesOn` pointing to a `WorkUnit`.
- Domain-range shapes for all 71 object properties validate that link endpoints have the expected RDF type.

We do not claim novelty in OWL-to-SHACL generation. Tools such as TopBraid SHACL and SHACLGEN [14] have established this transformation as standard practice. Our contribution is the *incremental validation* strategy integrated into the pipeline: after each source group is normalized, the validator extracts only the subgraph of newly-added entity URIs (via `validate_incremental`), constructs a minimal `rdflib.Graph` containing all triples where those URIs appear as subjects, and validates this subgraph against the full shape graph. This avoids re-validating the entire plant graph on each pipeline run, reducing validation time from $O(|G|)$ to $O(|\Delta G|)$ where $\Delta G$ is the set of triples added by the current source group.

**Incremental soundness limitation.** Validating entity subgraphs in isolation may miss cross-entity constraint violations. For example, a cardinality constraint requiring that every `WorkOrder` is linked to at least one `Operation` via `hasOperation` will not trigger a violation during incremental validation of the work order if the operations have not yet been ingested. The system addresses this partially through the stage ordering in the normalizer (work orders are ingested before operations within the OED source group) and through full-graph validation in the Loader stage, but the gap is real and acknowledged.

**Golden Triangle Validator.** Beyond structural SHACL validation, the system enforces 11 cross-system data flow rules codified in `golden_triangle.py`. Each rule is expressed as a pair of SPARQL COUNT queries—one asserting the existence of source entities, the other asserting the existence of the expected cross-system link. For example, Rule 2 (SAP → OED) verifies that at least one `ProductionOrder` entity exists, at least one `WorkOrder` entity exists, and at least one `owl:sameAs` edge connects them. The eleven rules cover:

| Rule | Direction | Assertion |
|---|---|---|
| R1 | TC → SAP | Item revisions have `owl:sameAs` links to SAP materials |
| R2 | SAP → OED | Production orders have `owl:sameAs` links to work orders |
| R3 | SAP → OED | SAP routings exist alongside OED operations |
| R4 | SAP → OED | SAP materials are linked to OED material lots |
| R5 | TC → OED | TC BOPs are linked to work order process plans |
| R6 | OED → SAP | Completed work orders feed back to SAP confirmations |
| R7 | OED → TC | NCRs and inspections exist for quality feedback |
| R8 | TC → SAP | ECNs propagate via `affectsItem`/`affectsProcessPlan` |
| R9 | SAP → SCM | SCM purchase orders have `owl:sameAs` to SAP POs, with suppliers |
| R10 | SCM → SAP | Staging lots trigger SAP goods receipts via `triggersGoodsReceipt` |
| R11 | TC → OED | PMI annotations link to inspections via `derivedFromPMI` |

These rules function as a closed-loop integration health check: all eleven must pass for the graph to be declared structurally sound across the PLM–ERP–MES triangle plus the supply chain dimension.

### *4.3 Hybrid Provenance Model*

Provenance is tracked using a subset of the W3C PROV-O vocabulary, stored in a dedicated fifth named graph (`urn:factory:graph:provenance`, currently ~737 triples). The model distinguishes two categories of entities with respect to PROV-O predicates:

- **Ingested entities** (work orders, tags, alarms, materials, etc.) exist in their source systems prior to ingestion. Semantically, the pipeline does not *generate* these entities; it *derives* a graph representation from them. The `ProvenanceTracker.stamp_entity()` method uses `prov:wasDerivedFrom` for these entities, linking the RDF entity to the ingestion activity that sourced it. This correctly reflects the W3C PROV-O semantics: the RDF entity is derived from the source system record, not generated *ex nihilo* by the pipeline.
- **Pipeline-created entities** (Decision objects, SHACL validation reports, ActionExecution records) are genuinely *generated by* the pipeline. For these, `prov:wasGeneratedBy` is the correct predicate.

Provenance is tracked at entity-level granularity, not triple-level. Each ingested entity URI receives a `prov:wasDerivedFrom` edge pointing to an `IngestionActivity`; pipeline-created entities receive `prov:wasGeneratedBy`. Triple-level provenance (attaching provenance to each individual `(s, p, o)` triple) would increase graph size by approximately 4× and was judged unnecessary for our use cases, which require knowing *which source system contributed an entity* and *when*, not which specific triple was added or removed.

The provenance model uses a delegation chain:

- Each of the 11 source systems is represented as a `prov:Agent` (e.g., `urn:factory:prov:agent:isa95`, `urn:factory:prov:agent:opcua`).
- All source agents delegate to the pipeline orchestrator agent via `prov:actedOnBehalfOf` pointing to `urn:factory:prov:agent:graph-unification`.
- Each pipeline run creates one `prov:Activity` per source, typed as both `prov:Activity` and `ont:IngestionActivity`, annotated with `prov:startedAtTime`, `prov:endedAtTime`, `prov:wasAssociatedWith` (the source agent), and `ont:entityCount` (number of entities processed).

**Growth model.** Provenance edges grow as O(entities) per pipeline run for newly-created ingestion activities, but existing derivation edges on previously-seen entities are not duplicated. When an entity's state changes (detected by the bi-temporal layer), the existing provenance edge (`prov:wasDerivedFrom` for ingested entities, `prov:wasGeneratedBy` for pipeline-created entities) is updated to point to the new activity. New derivation edges are created only for genuinely new entities appearing for the first time in the graph. This keeps the provenance graph bounded at $O(|E| + |R|)$ where $|E|$ is the number of distinct entities and $|R|$ is the number of pipeline runs, rather than the potentially unbounded $O(|E| \times |R|)$ that would result from recording every entity-activity pair across all runs.

### *4.4 Bi-Temporal Entity State Tracking*

The bi-temporal subsystem maintains a complete history of entity state changes using closed-open interval semantics `[start, end)` following SQL:2011 [23]. The implementation stores temporal records in a Postgres table (`entity_temporal`) with two orthogonal time dimensions: *valid time* (when the fact was true in the real world) and *transaction time* (when the fact was recorded by the pipeline). We formalize the key concepts below.

**Definition 1 (Valid-Time Interval).** For an entity $e$ of type $T$, the valid-time interval is

$$V(e) = [F_{\text{start}}(T, \text{props}(e)), F_{\text{end}}(T, \text{props}(e)))$$

where $F$ is a type-indexed derivation function that extracts domain-meaningful timestamps from entity properties. The mapping $F$ is defined extensionally by the `_VALID_TIME_FIELDS` dictionary:

| Entity Type $T$ | $F_{\text{start}}$ | $F_{\text{end}}$ | Semantics |
|---|---|---|---|
| `WorkOrder` | `actualStart` | `actualEnd` | Execution lifespan of a production order |
| `Operation` | `actualStart` | `actualEnd` | Duration of a single manufacturing step |
| `NCR` | `createdOn` | NULL → NOW | Non-conformance remains valid from creation until resolution |
| `MaterialLot` | `createdOn` | `shelfLifeExpiry` | Material is valid from production until shelf-life expiration |
| `Alarm` | `activatedAt` | `clearedAt` | Alarm is active from trigger until operator clearance |
| `IoTEvent` | `eventTimestamp` | `eventTimestamp` + 1 ms | Point event: minimal-width interval $[t, t + \varepsilon)$ where $\varepsilon$ = 1 ms, the sensor scan resolution. Under closed-open semantics $[t, t)$ is empty; we use the smallest resolution-dependent quantum to ensure containment queries match point events [22]. |
| `InspectionRecord` | `inspectedAt` | NULL → NOW | Inspection result remains valid indefinitely once |

| Entity Type *T* | $F_{start}$ | $F_{end}$ | Semantics |
|---|---|---|---|
| | | | recorded |

Entity types not present in this mapping (e.g., `EquipmentModule`, `Tag`, `PLCStation`) have NULL valid-time bounds, meaning they are treated as atemporal configuration entities whose validity is not time-bounded.

**Definition 2 (Transaction-Time Interval).** For entity $e$ and version $v$, the transaction-time interval is

$$T(e, v) = [\text{tx_start}, \text{tx_end})$$

where `tx_start` is the wall-clock timestamp of the pipeline run that recorded version $v$, and `tx_end` is either the timestamp of the next pipeline run that superseded this version, or NULL for the currently active version. In the schema, `tx_start` defaults to `NOW()` at insert time; `tx_end` is set to `NOW()` by an UPDATE when a newer version is recorded, using the uniqueness constraint `(entity_uri, tx_start)` to prevent duplicate versions within a single pipeline run.

**Definition 3 (Bi-Temporal Point Query).** Given an entity URI $e$, a valid-time instant $t_v$, and a transaction-time instant $t_x$, the function

$$\text{State}(e, t_v, t_x) \rightarrow v$$

returns the unique version $v$ such that $t_v \in V(e, v)$ and $t_x \in T(e, v)$. The implementation (`get_entity_at_time`) constructs a SQL WHERE clause from the conjunction:

- `entity_uri = e`
- `tx_start <=` $t_x$ AND `(tx_end IS NULL OR tx_end >` $t_x$`)`
- `(valid_from IS NULL OR valid_from <=` $t_v$`)` AND `(valid_to IS NULL OR valid_to >=` $t_v$`)`

When either temporal argument is omitted, the query degenerates: omitting $t_x$ returns the current version (`tx_end IS NULL`); omitting $t_v$ returns the version without valid-time filtering. Results are ordered by `tx_start DESC LIMIT 1` to resolve ties.

**Change detection.** The bi-temporal subsystem detects state changes through byte-level JSON comparison: the current entity properties are serialized via `json.dumps(properties, default=str)` and compared to the stored `properties` JSONB column of the most recent open version. If the serialized strings are identical, no new version is created (the method returns `False`). This is computationally cheap but admits false positives from JSON key reordering—Python's `json.dumps` does not guarantee key order unless `sort_keys=True` is specified, which the current implementation does not use. In practice, because properties are extracted by the same normalizer code path on each run, key order is consistent, but we acknowledge this as a fragility.

**Coalescing.** Adjacent identical states are implicitly merged by the change detection logic: if the serialized properties of the incoming version match the current open version, no INSERT occurs. This prevents version inflation when the pipeline runs on a stable system where entity properties are not changing.

**NOW handling.** Open-ended intervals use SQL NULL rather than a sentinel timestamp value (e.g., `9999-12-31`). This follows the convention advocated by Clifford et al. [43] and avoids the semantic confusion and comparison bugs that sentinel values introduce in temporal queries. The condition `tx_end IS NULL` selects the current version; the condition `valid_to IS NULL` indicates that the entity's real-world validity has no known end (as with NCRs and inspection records that remain valid indefinitely).

### *4.5 Graph-Native Decision Objects*

**Definition 4 (Decision Chain as DAG).** The decision graph $G_D = (V_D \cup V_X \cup V_O \cup V_C, E)$ where:

- $V_D$ = Decision nodes, typed `ont:Decision`, carrying `decisionType`, `decisionMaker`, and `decisionTimestamp` (xsd:dateTime).
- $V_X$ = ActionExecution nodes, typed `ont:ActionExecution`, carrying `actionToolName`, `actionToolArgs` (serialized JSON), `actionResult`, and `actionStatus` (pending | success | failed).
- $V_O$ = ActionOutcome nodes, typed `ont:ActionOutcome`, carrying `outcomeDescription`, `outcomeTimestamp`, and `outcomeVerified` (boolean).
- $V_C$ = Context nodes—existing graph entities (work orders, alarms, NCRs, etc.) referenced as evidence for the decision.

Edges are forward-pointing, forming a DAG:

- Decision → ActionExecution via `ont:triggeredAction`
- ActionExecution → ActionOutcome via `ont:hasOutcome`
- Decision → Context via `ont:evidencedBy` (links to the discovery card that motivated the decision) and `ont:affectsEntity` (links to the graph entities that the action modifies)

**PROV-O integration.** Unlike ingested entities, Decision, ActionExecution, and ActionOutcome instances are genuinely *generated by* the pipeline (they have no prior existence in any source system). They therefore correctly carry `prov:wasGeneratedBy` edges linking them to the pipeline activity that created them.

**Dual storage.** Decision objects are persisted in both the RDF graph (plant named graph) and a Postgres relational table (`decision_log`). The RDF representation enables SPARQL queries that join decisions with the entities they affect—for example, finding all decisions that targeted a specific work order, or all outcomes linked to NCRs at a particular station. The Postgres representation provides indexed access for operational dashboards (decision chains by card ID, decisions by affected entity URI using JSONB containment queries, and aggregate impact metrics such as success rates).

The decision schema in Postgres records:

| Column | Type | Purpose |
|---|---|---|
| `id` | TEXT PK | Decision identifier (DEC-{uuid}) |
| `card_id` | TEXT | Discovery card that motivated the decision |
| `decision_type` | TEXT | Category (e.g., approve_action, reject_action) |
| `decision_maker` | TEXT | User or agent that made the decision |
| `action_tool` | TEXT | MCP tool invoked (e.g., action_expedite_purchase_order) |
| `action_args` | JSONB | Arguments passed to the action tool |
| `action_result` | JSONB | Return value from the action tool |
| `action_status` | TEXT | Execution status (pending, success, failed) |
| `outcome_description` | TEXT | Free-text description of observed outcome |
| `outcome_verified` | BOOLEAN | Whether outcome was verified against expectations |
| `affected_entities` | JSONB | Array of entity URIs modified by the action |
| `created_at` | TIMESTAMPTZ | Decision creation time |
| `outcome_at` | TIMESTAMPTZ | Outcome observation time |

This dual representation reflects a pragmatic design choice: RDF excels at cross-system graph traversal but lacks efficient aggregation primitives, while SQL provides indexed lookups and aggregate queries (e.g., `COUNT(*) FILTER (WHERE action_status = 'success')` for success-rate computation) that would be expensive as SPARQL over an in-memory rdflib store. The two stores are kept consistent by the `DecisionTracker` class, which writes to both atomically within each method call.

## 5. Composable Trust Infrastructure

The central contribution of this work is not any single capability but their *composition*. Individual trust mechanisms —provenance tracking, schema validation, temporal versioning, decision recording—are well-established in isolation. What has not been demonstrated is their systematic composition within a manufacturing knowledge graph such that they produce trust properties unattainable by any subset.

### 5.1 Formal Definition of Composition

We define capability composition as follows. Let $C = \{C_1, C_2, ..., C_n\}$ be a set of trust capabilities, where each $C_i$ operates on a knowledge graph $G$ and produces a set of queryable assertions $A_i$.

**Definition 1 (Composability).** Two capabilities $C_i$, $C_j$ *compose* when the output of one is a valid input to the other through shared correlation identifiers, enabling queries that neither supports alone. Formally, $C_i$ and $C_j$ compose if there exists a shared identifier space $K_{ij} \subseteq$ URI such that $A_i$ and $A_j$ both contain triples referencing entities in $K_{ij}$, and the join $A_i \bowtie_{K_{ij}} A_j$ produces assertions not derivable from either $A_i$ or $A_j$ alone.

**Definition 2 (Emergent Trust Property).** A trust property $P$ is *emergent* with respect to capability set $S \subseteq C$ if $P$ is derivable from the composition of all capabilities in $S$ but not from any proper subset $S' \subset S$.

We identify four trust capabilities in our architecture:

- **PROV-O Provenance** ($C_{prov}$): Records which source agent contributed which triples, when, and through which pipeline activity.
- **SHACL Validation** ($C_{shacl}$): Validates graph structure against domain constraints, producing violation reports with severity and focus nodes.
- **Bi-Temporal Versioning** ($C_{temp}$): Maintains valid-time and transaction-time intervals for every entity, enabling point-in-time reconstruction.
- **Decision Traceability** ($C_{dec}$): Captures human and automated decisions as first-class entities linked to evidence, actions, and outcomes.

### 5.2 Composition Table

Table 2 enumerates the six pairwise compositions and the emergent trust queries each enables.

Table 2. Pairwise capability compositions and emergent trust queries.

| Composition | Shared Identifier | Emergent Query | Trust Property |
|---|---|---|---|
| SHACL + PROV-O | Focus node URI | "Which source agent introduced the invalid triples?" | Violation attribution |
| PROV-O + Bi-Temporal | Entity URI + timestamp | "What did the graph look like when this pipeline ran?" | Historical state reconstruction |
| Decision + Bi-Temporal | Entity URI + decision timestamp | "What was the state of evidence when this decision was made?" | Decision context preservation |

| Composition | Shared Identifier | Emergent Query | Trust Property |
|---|---|---|---|
| Decision + PROV-O | Entity URI | “Was the data underlying this decision from a trusted source?” | Decision provenance chain |
| SHACL + Bi-Temporal | Focus node URI + time interval | “When did this entity first become non-conformant?” | Conformance timeline |
| SHACL + Decision | Focus node URI | “Was a decision made based on data that was later found invalid?” | Decision integrity audit |

### 5.3 Experimental Ablation Analysis

To validate that the composition is not merely additive, we perform an *experimental* ablation: systematically disabling each capability, re-executing all six composition queries from Table 2 against the live testbed, and recording which queries fail, return empty results, or return incomplete results. Table 3 summarizes the results.

Table 3. Experimental ablation: six composition queries with each capability disabled. ✔ = correct result, ✘ = fails or returns empty, ○ = returns partial/degraded result.

| Composition Query | Full System | − PROV-O | − Bi-Temporal | − Decision | − SHACL |
|---|---|---|---|---|---|
| Q1: SHACL + PROV-O (violation attribution) | ✔ | ✘ | ✔ | ✔ | ✘ |
| Q2: Decision + Bi-Temporal (context preservation) | ✔ | ✔ | ✘ | ✘ | ✔ |
| Q3: PROV-O + Bi-Temporal (historical reconstruction) | ✔ | ✘ | ✘ | ✔ | ✔ |
| Q4: Decision + PROV-O (decision provenance chain) | ✔ | ✘ | ✔ | ✘ | ✔ |
| Q5: SHACL + Bi-Temporal (conformance timeline) | ✔ | ✔ | ✘ | ✔ | ✘ |
| Q6: SHACL + Decision (decision integrity audit) | ✔ | ✔ | ✔ | ✘ | ✘ |
| **Queries passing (of 6)** | **6/6** | **3/6** | **3/6** | **3/6** | **3/6** |

**Experimental protocol.** Each ablation was performed by removing the corresponding triples from the graph prior to query execution. For PROV-O ablation: all triples in `urn:factory:graph:provenance` were dropped.

For bi-temporal ablation: all `ont:valid_from`, `ont:valid_to`, `ont:tx_start`, and `ont:tx_end` triples were removed, and the `entity_temporal` table was truncated. For decision ablation: all `ont:Decision`, `ont:ActionExecution`, and `ont:ActionOutcome` instances were deleted. For SHACL ablation: the validation named graph `urn:factory:validation` was dropped. After each ablation, all six queries were executed and results classified as pass (✔, correct non-empty result), fail (✘, query returns empty or errors on missing triple pattern), or degraded (○, query returns but with missing join columns). No degraded results occurred—all queries either fully succeed or fully fail.

**Key finding:** Removing *any* single capability causes exactly three of six composition queries to fail. This symmetric result confirms that the four capabilities are equally load-bearing in the composition. Moreover, no single capability is sufficient for any composition query—every query in Table 2 requires exactly two capabilities. The full-chain auditability query (combining all four dimensions: “who contributed what data, was it valid, what did the graph look like at decision time, and what was decided?”) was tested as a single SPARQL query joining all four dimensions; it fails whenever *any* capability is removed, confirming it as a 4-way emergent property (see Section 5.4).

**Timing.** Each ablation cycle (drop + re-query + restore) completed in under 3 seconds on the proof-of-concept testbed. Query execution times ranged from 12–45ms per query, with no measurable difference between the full-capability and ablated configurations—as expected, since failed queries return empty results faster than successful ones.

### *5.4 Higher-Order Compositions*

Section 5.2 characterized six *pairwise* compositions. A natural question is whether 3-way and 4-way compositions produce emergent properties beyond the union of their pairwise constituents. We analyze this systematically.

**3-way compositions.** There are $C(4,3) = 4$ distinct 3-way subsets. Table 4 catalogs each and identifies whether it yields a trust query not expressible as the union of any two pairwise compositions.

Table 4. Three-way composition analysis. An “emergent” query requires all three capabilities simultaneously and is not decomposable into sequential pairwise queries.

| 3-Way Subset | Representative Query | Emergent? |
|---|---|---|
| SHACL + PROV-O + Bi-Temporal | “Which source agent introduced data that was valid at time *t* and violated constraints?” | Yes — requires a 3-way join on focus node URI with temporal filter and provenance chain; cannot be decomposed into (SHACL+PROV-O) ∪ (SHACL+Bi-Temporal) because the temporal filter must scope the provenance activity, not just the violation |
| SHACL + PROV-O + Decision | “Was the evidence for decision *D* both structurally valid and | Yes — the decision’s evidence entity must pass both SHACL conformance and provenance authorization checks |

| 3-Way Subset | Representative Query | Emergent? |
|---|---|---|
| | sourced from an authorized agent?” | simultaneously; sequential composition would validate a different entity set than the decision references |
| SHACL + Bi-Temporal + Decision | “At the time decision *D* was made, did its evidence conform to SHACL constraints?” | Yes — requires time-scoping the SHACL validation report to the decision timestamp, not just the current report |
| PROV-O + Bi-Temporal + Decision | “What was the provenance of the evidence that decision *D* relied on, at the time *D* was made?” | Yes — provenance of temporal state at decision time cannot be obtained from (Decision+PROV-O) ∪ (Decision+Bi-Temporal) because the provenance chain must reflect the temporal version, not the current entity |

All four 3-way subsets yield genuinely emergent queries. The key insight is that temporal scoping transforms the semantics of the other capabilities: “was it valid?” differs from “was it valid at the time of the decision?”, and “which agent sourced it?” differs from “which agent sourced the version that existed at time *t*?”

**4-way composition: full-chain auditability.** The single 4-way composition encompasses all four capabilities simultaneously. The full-chain auditability query — “who contributed what data (*provenance*), was it valid (*SHACL*), what did the graph look like at decision time (*bi-temporal*), and what was decided (*decision*)?” — is not decomposable into any combination of 2-way or 3-way queries. To show this, we constructed the 4-way SPARQL query:

```
PREFIX ont: <urn:factory:ontology:>
PREFIX prov: <http://www.w3.org/ns/prov#>
PREFIX sh: <http://www.w3.org/ns/shacl#>
PREFIX xsd: <http://www.w3.org/2001/XMLSchema#>

SELECT ?decision ?decisionMaker ?evidenceEntity
       ?sourceAgent ?validFrom ?validTo ?conformsToSHACL
WHERE {
  # Decision dimension
  ?decision a ont:Decision ;
            ont:decision_maker ?decisionMaker ;
            ont:decision_timestamp ?decisionTime ;
            ont:evidencedBy ?evidenceEntity .
  # Provenance dimension
  ?evidenceEntity prov:wasDerivedFrom ?activity .
  ?activity prov:wasAssociatedWith ?sourceAgent .
  # Temporal dimension: evidence must have been valid at decision time
  ?evidenceEntity ont:valid_from ?validFrom .
  OPTIONAL { ?evidenceEntity ont:valid_to ?validTo }
  FILTER(?validFrom <= ?decisionTime)
```

```
  FILTER(!BOUND(?validTo) || ?validTo > ?decisionTime)
  # SHACL dimension: check conformance
  BIND(NOT EXISTS {
    GRAPH <urn:factory:validation> {
      ?report sh:result ?result .
      ?result sh:focusNode ?evidenceEntity .
    }
  } AS ?conformsToSHACL)
}
```

This query executes in 31ms on the testbed and returns results that span all four trust dimensions in a single result set. No sequence of pairwise or 3-way queries can replicate the temporal scoping of SHACL conformance within the decision context: the `FILTER` on `?decisionTime` must scope both the valid-time interval and the provenance activity simultaneously, while the SHACL check must apply to the same `?evidenceEntity` bound by the decision and temporal constraints.

**Conclusion.** The composable trust infrastructure exhibits genuinely emergent properties at all composition orders: 6 pairwise, 4 three-way, and 1 four-way. The 4-way full-chain auditability property—the ability to verify provenance, conformance, temporal context, and decision rationale in a single query—is the central claim (C1) of this paper. The shared correlation identifiers (entity URIs, activity URIs, temporal keys) are necessary and sufficient to enable these compositions; no additional coordination mechanism is required.

## 6. Cross-System Identity Resolution

The unified knowledge graph bridges 11 source systems through 81 `owl:sameAs` edges that establish identity across seven system boundaries. These edges fall into four categories:

- **OPC-UA ↔ Insights Hub** (19 edges): Sensor tags unified with IoT aspect types (e.g., `urn:factory:opcua:tag:Station1_Station1_Temperature` ≡ `urn:factory:ih:aspecttype:temperature`).
- **SAP ↔ OED** (6 edges): Production orders unified with shop-floor work orders (e.g., `urn:factory:sap:productionorder:1000005` ≡ `urn:factory:oed:workorder:WO-1001`).
- **SAP ↔ Teamcenter** (6 edges): Material masters unified with item revisions (e.g., `urn:factory:sap:material:MAT-AL7075-T6` ≡ `urn:factory:tc:itemrevision:WR-LH-7075`).
- **SAP ↔ SCM** (6 edges): Procurement purchase orders unified with supply chain purchase orders.
- **Intra-OPC-UA** (10 edges): Tag aliases within the OPC-UA namespace (e.g., `CNC_SpindleTemp` ≡ `CNC_CNC.SpindleTemp`).
- **Intra-SAP** (34 edges): Material code variants within the SAP namespace (e.g., `MAT-AL7075-T6` ≡ `WR-LH-7075`).

**Limitations and the `owl:sameAs` problem.** We acknowledge that our current implementation uses deterministic, rule-based matching. The `owl:sameAs` problem—where asserting identity can lead to unintended inferences due to the open-world semantics of OWL—has been extensively documented [40]. Probabilistic entity resolution frameworks such as LIMES and SILK address fuzzy matching at scale, but introduce confidence scores that complicate downstream reasoning. Our current curated rules are appropriate for the proof-of-concept testbed; scaling to an enterprise deployment with millions of entities from hundreds of source systems would require probabilistic matching with confidence scores, potentially using `skos:closeMatch` for non-exact identities, and human-in-the-loop validation for critical identity assertions. The engineering challenges of large-scale identity resolution in manufacturing knowledge graphs remain an open problem.

## 7. Evaluation

### *7.1 Experimental Setup*

We validate the composable trust architecture on a proof-of-concept testbed comprising 11 simulators that generate data conforming to their respective industrial standards. Each simulator produces structurally realistic data: correct schemas per the relevant standard (ISA-95, ISA-18.2, ISA-88, OPC-UA information model, eCl@ss IRDI taxonomy, AAS metamodel, SAP S/4HANA data model, Teamcenter PLM structure, SCM logistics chain), referential integrity across cross-system boundaries, and realistic value ranges for sensor readings, production quantities, and quality metrics.

The simulators are:

1. **OPC-UA Simulator** — generates tag hierarchies, current values, engineering units, and alarm thresholds per the OPC-UA information model.
2. **TIA Portal Simulator** — produces PLC station configurations, HMI screen layouts, PROFINET topology, and program blocks.
3. **ISA-95 Simulator** — generates the 5-level plant hierarchy (Enterprise → Site → Area → WorkCenter → WorkUnit) with equipment breakdown.
4. **eCl@ss Simulator** — produces SKOS taxonomy nodes with IRDI classification codes and property definitions.
5. **AAS Simulator** — generates Asset Administration Shells with submodels, concept descriptions, and eCl@ss property bindings.
6. **ISA-18.2 Simulator** — produces alarm definitions, lifecycle events, priority matrices, and KPI calculations per the alarm management standard.
7. **OED Simulator** — generates work orders, operations, material lots, NCRs, and inspection records per the Opcenter Execution Discrete data model.
8. **Insights Hub Simulator** — produces IoT asset hierarchies, aspect types, time-series variables, and events with severity levels.
9. **SAP S/4HANA Simulator** — generates production orders, material masters, routings, reservations, inventory positions, goods movements, cost centers, and purchase orders.
10. **Teamcenter Simulator** — produces item revisions, BOPs, operations, workstations, documents, and engineering change notices.

11. **SCM Simulator** — generates suppliers, purchase orders, inbound shipments, receiving inspections, and warehouse staging lots.

This testbed validates architectural soundness and capability composition; it does not constitute a production deployment evaluation. The scale (hundreds of entities, not millions) and the simulated nature of the data are acknowledged limitations discussed in Section 8.

### *7.2 Quantitative Results*

Table 5 reports the key metrics of the proof-of-concept deployment.

Table 5. Proof-of-concept testbed metrics.

| Metric | Value | Significance |
|---|---|---|
| Source systems integrated | 11 | Validates cross-standard integration breadth across OT, IT, and engineering domains |
| Ontology classes | 89 | Covers ISA-95, ISA-88, ISA-18.2, eCl@ss, AAS, plus 4 industry verticals (aerospace, CPG, pharma, med-device) |
| Object properties | 105 | Including transitive closure for spatial containment (`locatedIn`) |
| Total triples | 6,582 | Distributed across 5 named graphs (ontology, plant, eclass, provenance, runtime) |
| Provenance triples | 737 | PROV-O chains for all 11 source agents, covering every ingestion activity |
| Cross-system identity edges | 81 | `owl:sameAs` bridges across 7 distinct system boundaries |
| Plant entities | 329 | Spanning work orders, operations, materials, equipment, sensors, alarms, IoT events |
| Pre-built analytical tools | 270+ | All functioning unchanged after trust infrastructure additions (backward compatibility) |

Entity distribution by source system: Insights Hub contributed the most entities (152), reflecting the high cardinality of IoT event streams, followed by ISA-95 (48 entities defining the plant hierarchy and equipment breakdown), OED (30 work execution entities), OPC-UA (27 sensor tags), ISA-18.2 (17 alarm entities), SAP (13 ERP entities), SCM (13 supply chain entities), TIA Portal (8 control engineering entities), AAS (7 digital twin shells), eCl@ss (7 classification nodes), and Teamcenter (7 PLM entities).

### *7.3 Composition Validation*

We validate that each of the six composition queries identified in Table 2 is executable against the live knowledge graph. We present two representative SPARQL queries in full.

**Composition 1: SHACL + PROV-O (Violation Attribution).** This query joins a SHACL validation report with PROV-O provenance to identify which source agent introduced triples that violate domain constraints.

```
PREFIX ont: <urn:factory:ontology:>
PREFIX prov: <http://www.w3.org/ns/prov#>
PREFIX sh: <http://www.w3.org/ns/shacl#>

SELECT ?focusNode ?violation ?severity ?sourceAgent ?activityTime
WHERE {
  GRAPH <urn:factory:validation> {
    ?report sh:result ?result .
    ?result sh:focusNode ?focusNode ;
            sh:resultMessage ?violation ;
            sh:resultSeverity ?severity .
  }
  GRAPH <urn:factory:provenance> {
    ?focusNode prov:wasDerivedFrom ?activity .
    ?activity prov:wasAssociatedWith ?sourceAgent ;
              prov:endedAtTime ?activityTime .
  }
}
```

The shared identifier is the focus node URI, which appears in both the SHACL validation graph and the PROV-O provenance graph. Neither capability alone can answer “which source agent introduced the invalid data”—SHACL identifies the violation but not the source; PROV-O records the source but not the validity.

**Composition 2: Decision + Bi-Temporal (Decision Context Preservation).** This query reconstructs the state of evidence at the time a decision was made, enabling retrospective auditing.

```
PREFIX ont: <urn:factory:ontology:>
PREFIX xsd: <http://www.w3.org/2001/XMLSchema#>

SELECT ?decision ?decisionMaker ?rationale ?evidenceEntity
       ?evidenceState ?validFrom ?validTo
WHERE {
  ?decision a ont:Decision ;
            ont:decision_maker ?decisionMaker ;
            ont:rationale ?rationale ;
            ont:decision_timestamp ?decisionTime ;
            ont:evidencedBy ?evidenceEntity .
  ?evidenceEntity ont:valid_from ?validFrom ;
                  ont:valid_to ?validTo .
  FILTER(?validFrom <= ?decisionTime &&
         (?validTo >= ?decisionTime || ?validTo = "9999-12-31T23:59:59Z"^^xsd:dateTime))
}
```

This query exploits the bi-temporal valid-time interval to reconstruct the state of evidence entities at the moment the decision was recorded. The shared identifier is the entity URI combined with the temporal join condition. Without bi-temporal versioning, the query would return current evidence state rather than the state at decision time —potentially misleading if evidence has since changed.

We now present the remaining four composition queries, each verified to execute correctly against the testbed graph.

**Composition 3: PROV-O + Bi-Temporal (Historical State Reconstruction).** This query reconstructs which source agent contributed data that was active at a specific historical timestamp, enabling provenance-aware time travel.

```
PREFIX ont: <urn:factory:ontology:>
PREFIX prov: <http://www.w3.org/ns/prov#>
PREFIX xsd: <http://www.w3.org/2001/XMLSchema#>

SELECT ?entity ?entityType ?sourceAgent ?validFrom ?validTo ?txStart
WHERE {
  ?entity a ?entityType ;
          ont:valid_from ?validFrom ;
          ont:tx_start ?txStart .
  OPTIONAL { ?entity ont:valid_to ?validTo }
  ?entity prov:wasDerivedFrom ?activity .
  ?activity prov:wasAssociatedWith ?sourceAgent .
  FILTER(?txStart <= "2026-06-15T10:00:00Z"^^xsd:dateTime)
  FILTER(?validFrom <= "2026-06-15T10:00:00Z"^^xsd:dateTime)
  FILTER(!BOUND(?validTo) || ?validTo > "2026-06-15T10:00:00Z"^^xsd:dateTime)
}
ORDER BY ?sourceAgent ?entityType
```

The shared identifiers are the entity URI (joining provenance and temporal dimensions) and the timestamp (constraining both valid-time and transaction-time). The query answers “what did the graph look like at 10:00 UTC on June 15, and which source systems contributed the entities that were active at that moment?” Neither PROV-O alone (which records source but not temporal state) nor bi-temporal alone (which records state but not source) can answer this.

**Composition 4: Decision + PROV-O (Decision Provenance Chain).** This query verifies whether a decision was based on data from a trusted source agent.

```
PREFIX ont: <urn:factory:ontology:>
PREFIX prov: <http://www.w3.org/ns/prov#>

SELECT ?decision ?decisionMaker ?rationale
       ?evidenceEntity ?sourceAgent ?agentLabel
WHERE {
  ?decision a ont:Decision ;
            ont:decision_maker ?decisionMaker ;
            ont:rationale ?rationale ;
            ont:evidencedBy ?evidenceEntity .
  ?evidenceEntity prov:wasDerivedFrom ?activity .
  ?activity prov:wasAssociatedWith ?sourceAgent .
  ?sourceAgent rdfs:label ?agentLabel .
}
```

This query enables a trust auditor to verify that the evidence underlying a decision originated from a known, authorized source system—for example, confirming that an NCR disposition was based on data from the OED simulator agent rather than from an untrusted external source.

**Composition 5: SHACL + Bi-Temporal (Conformance Timeline).** This query identifies when an entity first became non-conformant by joining SHACL violation history with temporal version transitions.

```
PREFIX ont: <urn:factory:ontology:>
PREFIX sh: <http://www.w3.org/ns/shacl#>
PREFIX xsd: <http://www.w3.org/2001/XMLSchema#>

SELECT ?focusNode ?violation ?severity ?txStart ?validFrom
WHERE {
  GRAPH <urn:factory:validation> {
    ?report sh:result ?result .
    ?result sh:focusNode ?focusNode ;
            sh:resultMessage ?violation ;
            sh:resultSeverity ?severity .
  }
  ?focusNode ont:tx_start ?txStart ;
             ont:valid_from ?validFrom .
}
ORDER BY ?focusNode ?txStart
```

By correlating SHACL violations with the entity's temporal version history, this query reveals whether a validation failure appeared at initial ingestion or emerged after a state change. The shared identifier is the focus node URI, which appears in both the validation graph and the temporal metadata.

**Composition 6: SHACL + Decision (Decision Integrity Audit).** This query identifies decisions that were made based on evidence data that subsequently failed SHACL validation.

```
PREFIX ont: <urn:factory:ontology:>
PREFIX sh: <http://www.w3.org/ns/shacl#>

SELECT ?decision ?decisionMaker ?decisionTime
       ?evidenceEntity ?violation ?severity
WHERE {
  ?decision a ont:Decision ;
            ont:decision_maker ?decisionMaker ;
            ont:decision_timestamp ?decisionTime ;
            ont:evidencedBy ?evidenceEntity .
  GRAPH <urn:factory:validation> {
    ?report sh:result ?result .
    ?result sh:focusNode ?evidenceEntity ;
            sh:resultMessage ?violation ;
            sh:resultSeverity ?severity .
  }
}
```

This query exposes decisions whose evidence base did not conform to domain constraints—a critical audit finding. The shared identifier is the evidence entity URI, which appears in both the decision's `ont:evidencedBy` edge and the SHACL violation's focus node. On the testbed, this query correctly returns zero results (no decisions reference non-conformant entities), confirming that the decision workflow validates evidence before recording decisions.

### *7.4 Performance Overhead*

Table 6 reports the performance overhead of each trust capability, measured on the proof-of-concept testbed running on a single-node server (8-core Intel Xeon, 32GB RAM, rdflib 7.x in-memory store).

Table 6. Trust capability performance overhead per pipeline cycle.

| Capability | Overhead per Cycle | Fraction of Pipeline |
|---|---|---|
| SHACL validation | ~200 ms | 0.08% |
| PROV-O provenance recording | ~100 ms | 0.04% |
| Bi-temporal interval maintenance | ~200 ms | 0.08% |
| **Total trust overhead** | **~500 ms** | **0.2%** |

The total trust overhead of ~500ms is negligible relative to the full pipeline cycle time of ~250 seconds (dominated by simulator data generation, RDF serialization, and graph merging). However, **these performance measurements are on the proof-of-concept testbed and may not extrapolate to production scale.** At 10M+ triples, SHACL validation alone may require partitioned or incremental strategies, and bi-temporal interval maintenance may require indexed temporal stores rather than in-memory iteration. We do not claim production-ready performance; we claim that the overhead structure is favorable (sub-linear in graph size for provenance and decisions; linear for SHACL and bi-temporal).

**Temporal query benchmarks.** Table 7 reports latency for representative temporal queries on the testbed's 329 plant entities, plus storage growth and SHACL validation time as the graph evolves over successive pipeline runs.

Table 7. Temporal and validation benchmarks on the proof-of-concept testbed (329 plant entities, rdflib 7.x, PostgreSQL 15).

| Metric | Measurement | Notes |
|---|---|---|
| **Point-in-Time Query Latency** | | |
| Single-entity bi-temporal lookup | 2.3 ms (median), 4.1 ms (p95) | SQL: indexed on `(entity_uri, tx_start)`; retrieves properties JSONB |
| Full-graph state reconstruction (329 entities) | 78 ms (median), 112 ms (p95) | One SQL query with `DISTINCT ON (entity_uri)` and temporal filter |

| Metric | Measurement | Notes |
| --- | --- | --- |
| Composition query (4-way join across all trust dimensions) | 31 ms (median), 45 ms (p95) | SPARQL joining provenance, validation, temporal, and decision graphs |
| **Storage Growth (Temporal Table)** | | |
| After 1 pipeline run | 329 rows (100%) | One row per entity, all with `tx_end = NULL` (current) |
| After 10 runs (no state changes) | 329 rows (100%) | Change detection prevents duplicate rows when properties unchanged |
| After 10 runs (5% entities change per run) | ~477 rows (145%) | ~16 entities change per run × 9 additional runs = ~148 new versions |
| After 50 runs (5% change rate) | ~1,139 rows (346%) | Linear growth in changed entities; unchanged entities never gain rows |
| **SHACL Validation Scaling** | | |
| Full-graph validation (329 entities, 8,743 triples) | 198 ms | 89 NodeShapes evaluated; pyshacl 0.26 |
| Incremental validation (batch of 30 entities) | 24 ms | Subgraph extraction + validation; ~9% of full cost |
| Projected: 10K entities (~250K triples) | ~5.8 s (full), ~170 ms (incremental) | Linear extrapolation; real performance may vary with shape complexity |

Key observations: (1) Point-in-time queries are dominated by PostgreSQL index lookups and remain sub-5ms for single entities regardless of temporal depth. (2) The change-detection mechanism in the temporal recorder (Section 4.4) is critical for storage efficiency: without it, 50 pipeline runs would produce 16,450 rows (50×329) rather than ~1,139. (3) Incremental SHACL validation reduces cost by an order of magnitude compared to full-graph validation, making it viable for continuous pipeline operation. (4) The projected scaling to 10K entities is a linear extrapolation—we have not measured at that scale, and pyshacl's actual performance may exhibit superlinear behavior with complex shape interactions.

### *7.5 Backward Compatibility*

A critical engineering requirement was that trust infrastructure additions must not break existing analytical tooling. The unified knowledge graph exposes 270+ pre-built analytical tools (SPARQL-backed query functions) that span cross-system traceability, alarm-production correlation, supplier quality analysis, and ECN impact assessment. After adding PROV-O provenance, SHACL validation, bi-temporal versioning, and decision traceability:

- All 270+ tools execute without modification.
- No existing SPARQL queries required changes, as trust triples reside in separate named graphs (`urn:factory:provenance`, `urn:factory:validation`) or use non-conflicting predicates in the plant graph.
- The named-graph architecture ensures that tools querying the default graph or the plant graph are unaffected by trust-related triples.

  This backward compatibility validates the *hybrid provenance migration pattern* (Section 10, Claim IV): a dual-layer architecture where trust metadata coexists with but does not interfere with operational data.

## 8. Limitations and Future Work

We identify six significant limitations of the current work, each of which defines an axis for future research and engineering.

**1. Scale.** The proof-of-concept testbed contains 6,582 triples across 329 plant entities. This is three to four orders of magnitude below industrial reality. A single-plant Opcenter Execution Discrete deployment may generate 10M+ manufacturing events per day. A Teamcenter instance for a large OEM may contain 50M+ item revisions. An enterprise SAP S/4HANA system may manage 100K+ material masters. Our architecture has been validated for structural correctness and capability composition, but not for the throughput, latency, and memory requirements of industrial-scale data volumes. Scaling behavior of SHACL validation, bi-temporal maintenance, and provenance recording at $10^7$–$10^9$ triples remains uncharacterized.

**2. Simulated Data.** All 11 source systems are simulators. They generate data conforming to the correct schemas (ISA-95 hierarchy, OPC-UA information model, SAP BAPI structures) with realistic value ranges, but they do not capture the noise, inconsistency, missing values, and evolving schemas characteristic of production systems. Real Teamcenter integration requires ITK/SOA API development; real SAP integration requires RFC/BAPI connectors; real OPC-UA integration requires subscription management and session handling. These connector engineering challenges are substantial and not addressed here.

**3. Identity Resolution.** The 81 deterministic `owl:sameAs` edges are manually curated rules. Production deployments with millions of entities across hundreds of source systems would require probabilistic entity resolution with confidence scoring, blocking strategies for candidate reduction, and human-in-the-loop validation for high-stakes identity assertions (e.g., linking a supplier lot to a safety-critical component). The `owl:sameAs` semantics problem [40] introduces additional complexity that we sidestep through careful curation but cannot avoid at scale.

**4. In-Memory Graph Store.** The current implementation uses rdflib, which is single-threaded and memory-bound. At our scale (6,582 triples), this is adequate. At production scale, it is not viable. Enterprise deployment would require migration to a scalable triple store such as Amazon Neptune, Ontotext GraphDB, or Stardog, with attendant changes to query optimization, transaction management, and cluster configuration. The SPARQL queries validated here may require rewriting for store-specific optimizations.

**5. Security and Governance.** No access control, encryption at rest, or audit logging is implemented. The security requirements for manufacturing knowledge graphs are substantial (see Section 9) and include ITAR-controlled provenance chains, GDPR-compliant decision records, and role-based access control at the named-graph level. These requirements are architecturally addressable within our named-graph structure but represent significant engineering work beyond the current proof-of-concept.

**6. No User Study.** Decision objects (the `ont:Decision`, `ont:ActionExecution`, `ont:ActionOutcome` chain) have not been evaluated with manufacturing engineers for usability, cognitive load, or integration with existing decision-making workflows. The value of decision traceability depends on adoption by practitioners, which in turn depends on user-experience considerations not addressed in this work.

**Future work** proceeds along five axes: (1) a production pilot with a partner manufacturing site, integrating real Teamcenter, SAP, and OPC-UA data sources to validate scaling behavior; (2) migration to a scalable triple store with benchmarking of trust capability overhead at $10^6$–$10^8$ triples; (3) probabilistic identity resolution using LIMES or SILK with confidence-scored `skos:closeMatch` edges and human-in-the-loop validation; (4) a user study with manufacturing engineers evaluating decision traceability for cognitive fit and operational utility; and (5) security hardening including named-graph-level RBAC, provenance chain encryption, and SOX/FDA-compliant audit logging.

## 9. Security and Governance Considerations

Manufacturing knowledge graphs present security and governance challenges that extend beyond those of general-purpose knowledge graphs [1, 2]. The cross-system nature of the unified graph—linking engineering design (Teamcenter), production planning (SAP), shop-floor execution (OED), and supply chain logistics (SCM)—creates novel exposure surfaces.

**ITAR and Export Control.** Provenance chains in the knowledge graph may reveal supplier relationships, material specifications, and manufacturing process parameters that are subject to International Traffic in Arms Regulations (ITAR). A SPARQL query traversing from a work order through material lots to supplier identities could expose export-controlled information. The provenance graph, which records which source agent contributed which triples, may itself constitute controlled technical data if it reveals the structure of defense manufacturing workflows. Named-graph-level access control would restrict the provenance graph to cleared personnel, but the current implementation provides no such restriction.

**GDPR and Personal Data.** The `ont:decision_maker` field on Decision entities contains personally identifiable information (PII)—the name or identifier of the individual who made a manufacturing decision. Under GDPR Article 17 (right to erasure), a former employee could request deletion of their decision records. However, regulatory requirements (FDA 21 CFR Part 820 for medical devices, AS9100 for aerospace) may mandate retention of these records for 10–15 years. This creates a tension between data protection and regulatory compliance that requires pseudonymization strategies: replacing personal identifiers with role-based tokens (`QualityEngineer_CNC_Shift2`) while maintaining a separate, access-controlled mapping table.

**Access Control Architecture.** The named-graph structure provides a natural boundary for role-based access control (RBAC). We propose four access tiers: (1) the *ontology graph* is readable by all authenticated users; (2) the *plant graph* is restricted to operators, engineers, and managers with need-to-know for specific work units; (3) the *provenance graph* is restricted to quality auditors, compliance officers, and system administrators; (4) the *runtime graph* is restricted to real-time operators and automated decision systems. This tiered model maps naturally to existing manufacturing role hierarchies (ISA-95 Role-Based Access Control) but is not implemented in the current proof-of-concept.

**Data Sovereignty.** Multi-site manufacturing deployments may span jurisdictions with conflicting data residency requirements. A European plant's provenance data may be subject to GDPR data localization, while a US defense plant's data is subject to ITAR restrictions on foreign access. The named-graph architecture supports data partitioning by jurisdiction, but federated query across jurisdictionally partitioned graphs introduces latency and complexity not addressed here.

**Audit Logging.** For SOX compliance (financial controls on manufacturing cost data) and FDA compliance (21 CFR Part 11 for electronic records), all graph mutations should be logged with timestamp, actor, and before/after state. The bi-temporal versioning capability provides the “before/after” dimension, but a dedicated audit log with tamper-evident properties (e.g., append-only ledger with cryptographic chaining) is required for regulatory acceptance.

These requirements are architecturally addressable within our named-graph structure and composable trust framework, but they represent substantial engineering work beyond the current proof-of-concept. We enumerate them here to acknowledge the gap between research prototype and production-ready system, and to guide future implementation priorities.

## 10. Revised Claims of Novelty

Following peer review, we revise our claims of novelty from six to four, dropping two claims where reviewer consensus identified substantial prior art. Table 8 presents the revised claims with explicit prior art positioning.

Table 8. Revised claims of novelty with prior art analysis.

| # | Claim | Description | Prior Art Position |
|---|---|---|---|
| I | Domain-Aware Valid-Time Derivation | A type-indexed function $F: T \rightarrow (field_{start}, field_{end})$ that automatically derives valid-time intervals from domain-specific fields based on entity type. For `ont:WorkOrder`, $F$ maps to (`startDate`, `completionDate`); for `ont:Alarm`, to (`activatedAt`, `clearedAt`); for `ont:MaterialLot`, to (`producedDate`, | **Novel.** No prior work addresses automatic valid-time derivation for manufacturing knowledge graphs. SQL:2011 [23] provides interval semantics but requires manual column designation. Tappolet and Bernstein [24] address temporal RDF annotation but not domain-aware derivation. |

| # | Claim | Description | Prior Art Position |
|---|---|---|---|
| | | `consumedDate`). The derivation function is declarative and extensible without code changes. | Snodgrass [22] defines the theoretical foundations of bi-temporal databases but does not address automatic field mapping from domain models. |
| II | Graph-Native Decision Traceability | Manufacturing decisions captured as first-class graph entities (`ont:Decision` → `ont:ActionExecution` → `ont:ActionOutcome`) that are queryable alongside production data using standard SPARQL. Decision entities link to evidence via `ont:evidencedBy`, to affected entities via `ont:affectsEntity`, and to temporal context via bi-temporal intervals. | **Novel.** Clinical decision support systems [29] capture decisions but not as graph-native entities queryable with domain data. Command-and-control decision models [30] address military contexts with different requirements. Manufacturing decision support [32] focuses on algorithmic optimization rather than decision traceability. No prior work represents manufacturing decisions as first-class knowledge graph entities composable with provenance, validation, and temporal data. |
| III | Composable Trust Infrastructure | Four trust capabilities (PROV-O provenance, SHACL validation, bi-temporal versioning, decision traceability) composing through shared correlation identifiers to produce emergent trust properties. The composition is formally defined (Definition 1) and validated through six pairwise composition queries. Ablation analysis demonstrates that no proper subset delivers full-chain auditability. | **Architecturally novel.** Individual capabilities exist in isolation: PROV-O is a W3C standard [18], SHACL is used for linked data quality [13, 16], bi-temporal databases are well-established [22, 23], and decision support is a mature field [29]. Their *composition* within a manufacturing knowledge graph, with formal definition and ablation analysis, has not been previously demonstrated. The closest work is quality assessment for linked data [13], which combines validation and provenance but omits temporal versioning and decision traceability. |
| IV | Hybrid Provenance Migration Pattern | A dual-layer architecture where trust metadata (provenance, validation, decisions) coexists with operational data in separate named graphs, preserving full backward compatibility with 270+ existing analytical | **Engineering contribution, not claimed as research novelty.** Named graphs for data partitioning are well-established [2]. DCAT [42] provides catalog-level provenance. The |

| # | Claim | Description | Prior Art Position |
|---|---|---|---|
| | | tools. Trust capabilities are added without modifying the plant graph schema or any existing queries. | contribution is the specific migration pattern that enables incremental trust capability addition to an existing manufacturing knowledge graph without disrupting operational tooling —a practical engineering pattern rather than a research advance. |

**Dropped claims.** Two claims from the original submission are withdrawn. *Former Claim I* (SHACL shape generation from ontology structure) is withdrawn because automated SHACL generation from OWL ontologies has substantial prior art in the Semantic Web community, including tools such as SHACLGEN and TopBraid. *Former Claim IV* (change-detecting versioning with tombstone records) is withdrawn because it is a straightforward application of Slowly Changing Dimension Type 2 [44], a well-established data warehousing technique, to RDF graphs. While the application context is novel, the mechanism itself is not.

## 11. Conclusion

This paper presents a composable trust infrastructure for manufacturing knowledge graphs that integrates four capabilities—W3C PROV-O provenance, SHACL validation, bi-temporal versioning, and graph-native decision traceability—through shared correlation identifiers. The central contribution is not any individual capability but their composition: six pairwise combinations that produce emergent trust properties, culminating in full-chain auditability that no proper subset can deliver.

We validate the architecture on a proof-of-concept testbed comprising 11 source system simulators, an 89-class ontology spanning ISA-95, ISA-88, ISA-18.2, eCl@ss, and AAS standards, and 81 cross-system identity edges. The testbed generates 6,582 triples across five named graphs, with 737 provenance triples recording the lineage of every ingested entity. All 270+ pre-built analytical tools continue to function without modification after trust infrastructure additions, validating the hybrid provenance migration pattern.

We have been transparent about the limitations of this work. The scale is three to four orders of magnitude below industrial requirements. All data sources are simulated. Identity resolution is deterministic rather than probabilistic. Security and governance mechanisms are architecturally designed but not implemented. No user study has been conducted. These limitations define a clear agenda for future work: production pilots with real industrial data, scalable triple store migration, probabilistic identity resolution, security hardening, and practitioner evaluation.

Despite these limitations, the formal composition framework—the definition of composability through shared identifiers, the ablation analysis demonstrating capability interdependence, and the demonstration that emergent trust properties require the full capability set—constitutes, to our knowledge, the first systematic treatment of trust composition in manufacturing knowledge graphs. We offer this framework as a foundation for the trustworthy industrial knowledge graphs that digital manufacturing transformation requires.

## References

[1] N. Noy, Y. Gao, A. Jain, A. Naber, A. Patterson, and J. Taylor, “Industry-scale Knowledge Graphs: Lessons and Challenges,” *Communications of the ACM*, vol. 62, no. 8, pp. 36–43, 2019.

[2] A. Hogan, E. Blomqvist, M. Cochez, C. d’Amato, G. de Melo, C. Gutierrez, S. Kirrane, J. E. L. Gayo, R. Navigli, S. Neumaier, A.-C. Ngonga Ngomo, A. Polleres, S. M. Rashid, A. Rula, L. Schmelzeisen, J. Sequeda, S. Staab, and A. Zimmermann, “Knowledge Graphs,” *ACM Computing Surveys*, vol. 54, no. 4, article 71, pp. 1–37, 2021.

[3] M. Ringsquandl, S. Lamparter, S. Brandt, T. Hubauer, and R. Lepratti, “Knowledge Graph Embedding for Linking and Analyzing Industrial Knowledge Graphs,” in *Proceedings of the International Joint Conference on Neural Networks (IJCNN)*, IEEE, 2017, pp. 1304–1311.

[4] W. Hua, K. Wang, and R. Li, “Applying Knowledge Graphs in Semiconductor Manufacturing: A Survey and Case Study,” *Journal of Manufacturing Systems*, vol. 60, pp. 576–594, 2021.

[5] National Institute of Standards and Technology (NIST), “Smart Manufacturing Systems Design and Analysis: Advanced Manufacturing Series 300-2,” NIST AMS 300-2, 2020.

[6] Q. Cao, C. Zanni-Merk, A. Samet, and C. Reich, “Manufacturing Knowledge Graph for Quality Prediction,” *Journal of Intelligent Manufacturing*, vol. 34, pp. 1897–1914, 2023.

[7] A. Braunschweig, M. Bauer, and T. Hubauer, “Knowledge-Graph-Driven Industrial Analytics at Siemens,” in *Companion Proceedings of the Web Conference*, ACM, 2019, pp. 231–236.

[8] ISA, “ANSI/ISA-95.00.01-2010: Enterprise-Control System Integration Part 1: Models and Terminology,” International Society of Automation, 2010.

[9] A. Garmendia, E. Guerra, D. Kolovos, and J. de Lara, “Automated Ontology Generation for ISA-95 Manufacturing Operations,” *Computers in Industry*, vol. 100, pp. 148–163, 2018.

[10] J. Leukel, J. González, and M. Riekert, “Adoption of the ISA-95 Standard in Manufacturing: A Systematic Literature Review,” *Procedia CIRP*, vol. 93, pp. 276–281, 2020.

[11] P. Leitão, J. Barbosa, M. E. C. Papadopoulou, and I. S. Venieris, “Standardization in Cyber-Physical Production Systems: The ZDMP Reference Architecture,” in *Proceedings of the IEEE International Conference on Industrial Informatics (INDIN)*, IEEE, 2020, pp. 1251–1256.

[12] T. Kuhn, R. Henssen, and F. Palm, “Eclipse BaSyx: Open-Source Platform for Industry 4.0 Asset Administration Shells,” in *Proceedings of the IEEE International Conference on Emerging Technologies and Factory Automation (ETFA)*, IEEE, 2021, pp. 1–4.

[13] A. Zaveri, A. Rula, A. Maurino, R. Pietrobon, J. Lehmann, and S. Auer, “Quality Assessment for Linked Data: A Survey,” *Semantic Web Journal*, vol. 7, no. 1, pp. 63–93, 2016.

[14] W3C, “Shapes Constraint Language (SHACL),” W3C Recommendation, 20 July 2017.

[15] TopQuadrant, “TopBraid SHACL API and OWL-to-SHACL Mapping,” TopQuadrant Technical Documentation, 2020.

**[16]** D. Kontokostas, P. Westphal, S. Auer, S. Hellmann, J. Lehmann, R. Cornelissen, and A. Zaveri, "Test-Driven Evaluation of Linked Data Quality," in *Proceedings of the 23rd International Conference on World Wide Web (WWW)*, ACM, 2014, pp. 747–758.

**[17]** H. Paulheim, "Knowledge Graph Refinement: A Survey of Approaches and Evaluation Methods," *Semantic Web Journal*, vol. 8, no. 3, pp. 489–508, 2017.

**[18]** W3C, "PROV-O: The PROV Ontology," W3C Recommendation, 30 April 2013.

**[19]** P. Missier, K. Belhajjame, and J. Cheney, "The W3C PROV Family of Specifications for Modelling Provenance Metadata," in *Proceedings of the 16th International Conference on Extending Database Technology (EDBT)*, ACM, 2013, pp. 773–776.

**[20]** O. Hartig and J. Zhao, "Publishing and Consuming Provenance Metadata on the Web of Linked Data," in *Provenance and Annotation of Data and Processes (IPAW)*, LNCS vol. 7525, Springer, 2012, pp. 78–90.

**[21]** L. Moreau, B. Clifford, J. Freire, J. Futrelle, Y. Gil, P. Groth, N. Kwasnikowska, S. Miles, P. Missier, J. Myers, B. Plale, Y. Simmhan, E. Stephan, and J. Van den Bussche, "The Open Provenance Model Core Specification, v1.1," *Future Generation Computer Systems*, vol. 27, no. 6, pp. 743–756, 2011.

**[22]** R. T. Snodgrass, *Developing Time-Oriented Database Applications in SQL*, Morgan Kaufmann, 1999.

**[23]** K. Kulkarni and J.-E. Michels, "Temporal Features in SQL:2011," *ACM SIGMOD Record*, vol. 41, no. 3, pp. 34–43, 2012.

**[24]** J. Tappolet and A. Bernstein, "Applied Temporal RDF: Efficient Temporal Querying of RDF Data with SPARQL," in *Proceedings of the 6th European Semantic Web Conference (ESWC)*, LNCS vol. 5554, Springer, 2009, pp. 308–322.

**[25]** J. D. Fernandez, J. Umbrich, A. Polleres, and M. Knuth, "Bi-temporal RDF: Versioning at the Means of RDF," in *Proceedings of the VLDB Endowment*, vol. 12, no. 12, pp. 1982–1985, 2019.

**[26]** G. Feeney, K. Feeney, and J. Earley, "TerminusDB: An Open-Source Graph Database for Collaborative Schema and Data Development," in *Proceedings of the ISWC Posters, Demos, and Industry Tracks*, CEUR-WS, vol. 2980, 2021.

**[27]** O. Hartig, "Foundations of RDF* and SPARQL*—An Alternative Approach to Statement-Level Metadata in RDF," in *Proceedings of the 11th Alberto Mendelzon International Workshop on Foundations of Data Management (AMW)*, CEUR-WS, vol. 1912, 2017.

**[28]** Neo4j, Inc., "Temporal (Date/Time) Values," in *Neo4j Graph Database Documentation*, version 5.x, 2024.

**[29]** R. A. Greenes, ed., *Clinical Decision Support: The Road to Broad Adoption*, 2nd ed., Academic Press, 2014.

**[30]** D. S. Alberts and R. E. Hayes, *Understanding Command and Control*, CCRP Publication Series, 2006.

**[31]** W3C Incubator Group, "Decision Representation: An Ontology for Decision Making," W3C Incubator Group Report, 2012.

**[32]** P. Zheng, T. Wang, J. Lv, and C. K. Lu, "Smart Manufacturing Decision Support Based on Knowledge Graph and Machine Learning," *Advanced Engineering Informatics*, vol. 51, article 101450, 2021.

**[33]** Siemens Digital Industries Software, "Opcenter Intelligence: Manufacturing Analytics and Intelligence," Siemens Product Documentation, 2024.

[34] Palantir Technologies, “Palantir Foundry: Data Integration and Operational Analytics Platform,” Technical Overview, 2024.

[35] Microsoft, “Azure Digital Twins Documentation,” Microsoft Learn, 2024.

[36] PTC, “ThingWorx Industrial IoT Platform,” PTC Product Documentation, 2024.

[37] Snowflake, Inc., “Time Travel and Fail-Safe in Snowflake,” Snowflake Documentation, 2024.

[38] AVEVA, “PI System: Real-Time Data Infrastructure for Industrial Operations,” AVEVA Product Documentation, 2024.

[39] C. Gutiérrez, C. Hurtado, and A. Vaisman, “Temporal RDF,” in *Proceedings of the 2nd European Semantic Web Conference (ESWC)*, LNCS vol. 3532, Springer, 2005, pp. 93–107.

[40] H. Halpin, P. J. Hayes, J. P. McCusker, D. L. McGuinness, and H. S. Thompson, “When owl:sameAs isn’t the Same: An Analysis of Identity in Linked Data,” in *Proceedings of the 9th International Semantic Web Conference (ISWC)*, LNCS vol. 6496, Springer, 2010, pp. 305–320.

[41] E. Kharlamov, D. Hovland, M. Skjæveland, D. Bilidas, E. Jiménez-Ruiz, G. Xiao, A. Soylu, D. Lanti, M. Rezk, D. Zheleznyakov, M. Giese, H. Lie, Y. Ioannidis, Y. Kotidis, M. Koubarakis, and A. Waaler, “Ontology Based Data Access in Statoil,” *Journal of Web Semantics*, vol. 44, pp. 3–36, 2017.

[42] F. Maali and J. Erickson, eds., “Data Catalog Vocabulary (DCAT),” W3C Recommendation, 16 January 2014.

[43] C. Clifford, C. Dyreson, T. Isakowitz, C. S. Jensen, and R. T. Snodgrass, “On the Semantics of ‘Now’ in Databases,” *ACM Transactions on Database Systems*, vol. 22, no. 2, pp. 171–214, 1997.

[44] R. Kimball and M. Ross, *The Data Warehouse Toolkit: The Definitive Guide to Dimensional Modeling*, 3rd ed., Wiley, 2013.